\documentclass[runningheads]{llncs}
\usepackage{graphicx}
\usepackage{hyperref}
\usepackage{latexsym}
\usepackage{url}
\usepackage{xcolor}
\usepackage{amssymb}
\usepackage{amsmath}
\usepackage[nolist]{acronym}
\usepackage{rotating}
\definecolor{newcolor}{rgb}{.8,.349,.1}

\acrodef{LSCC}[LSCC]{lung squamous cell carcinoma}
\acrodef{LUAD}[LUAD]{lung adenocarcinoma}
\acrodef{MEN}[MEN]{meningioma}
\acrodef{TCIA}[TCIA]{The Cancer Imaging Archive}
\acrodef{TCGA}[TCGA]{The Cancer Genome Atlas}
\acrodef{SOTA}[SOTA]{state-of-the-art}
\acrodef{WSI}[WSI]{whole slide image}
\acrodef{HE}[H\&E]{hematoxylin and eosin}
\acrodef{MIL}[MIL]{multiple instance learning}
\acrodef{H&E}[H&E]{hematoxilin and eosin}
\acrodef{DL}[DL]{deep learning}
\acrodef{SSL}[SSL]{self supervised learning}
\acrodef{OD}[OD]{optical density}
\acrodef{mpp}[mpp]{microns per pixel}
\acrodef{PHI}[PHI]{protected health information}

\begin{document}

\title{Re-identification from histopathology images}

\author{Jonathan Ganz\inst{1} \and
Jonas Ammeling\inst{1} \and
Samir Jabari\inst{2} \and
Katharina Breininger\inst{3} \and
Marc Aubreville\inst{1}}

% \authorrunning{J. Ganz et al.}

\institute{Technische Hochschule Ingolstadt, Ingolstadt, Germany \and
Institut für Neuropathologie, Universitätsklinikum Erlangen, Friedrich-Alexander-Universität Erlangen-Nürnberg \and
Department Artificial Intelligence in Biomedical Engineering, Friedrich-Alexander-Universität Erlangen-Nürnberg}

\maketitle
\begin{abstract}
In numerous studies, deep learning algorithms have proven their potential for the analysis of histopathology images, for example, for revealing the subtypes of tumors or the primary origin of metastases. These models require large datasets for training, which must be anonymized to prevent possible patient identity leaks. This study demonstrates that even relatively simple deep learning algorithms can re-identify patients in large histopathology datasets with substantial accuracy. We evaluated our algorithms on two TCIA datasets including lung squamous cell carcinoma (LSCC) and lung adenocarcinoma (LUAD). We also demonstrate the algorithm's performance on an in-house dataset of meningioma tissue. We predicted the source patient of a slide with $F_1$ scores of $50.16 \%$ and $52.30 \%$ on the LSCC and LUAD datasets, respectively, and with $62.31 \%$ on our meningioma dataset. Based on our findings, we formulated a risk assessment scheme to estimate the risk to the patient's privacy prior to publication.
\end{abstract}
%%Graphical abstract
% \begin{graphicalabstract}
% \includegraphics{grabs}
% \end{graphicalabstract}

%%Research highlights
% \begin{highlights}
% \item Research highlight 1
% \item Research highlight 2
% \end{highlights}

% \begin{keyword}
% %% keywords here, in the form: keyword \sep keyword
% digital pathology \sep deep learning \sep re-identification
% %% PACS codes here, in the form: \PACS code \sep code
% %\PACS 0000 \sep 1111
% %% MSC codes here, in the form: \MSC code \sep code
% %% or \MSC[2008] code \sep code (2000 is the default)
% %\MSC 0000 \sep 1111
% \end{keyword}

% \end{frontmatter}

%% \linenumbers

%% main text
\section{Introduction}
% -> What are whole slide images are used for?
% -> routine pathology, examination of slides by a pathologist is still gold standard
% -> Advent of digital pathology and automatic assessment of tissue specimens with DL
% -> Multiple examples of DL algorithms on histological slides
% -> Prediction of information that is not percivable by humans like gentic alterations
The evaluation of \ac{HE}-stained sections under the light microscope is a standard procedure in the pathological diagnosis of tumors. With the advent of whole slide scanners, it became possible to digitize glass slides into so-called \acp{WSI}. This allowed the use of automatic methods for tumor assessment, and in particular \ac{DL} algorithms have revolutionized the field of digital histopathology, as these models have shown promising performance in a wide variety of tasks such as automatic tumor grading and classification \cite{NIR2018167,han2017breast,ganz2021} or the automatic assessment of biomarkers including mitotic count \cite{aubreville2023mitosis,veta2019predicting} or the segmentation of tumor area \cite{Wilm2022}. Some \ac{DL} algorithms are even able to retrieve information from \acp{WSI} that is hidden from human experts like the prediction of molecular alterations \cite{coudray2018classification,hong2021predicting,lu2021clam} or the prediction of the primary origin of metastases \cite{lu2021cup}.
% -> Recently shown potential of DL to identify various information from histopathological slides
% -> cite CLAM/ cancers of unknown origin/
% -> Transition to for what re-identification could be used
% -> DL - Algorithms rely on large amounts of data

The majority of \ac{DL} approaches have in common that they need immense amounts of data to be trained on. This led to the publication of large-scale histopathology datasets to accelerate \ac{DL} research like the CAMELYON dataset \cite{litjens20181399}, the Breast Invasive Carcinoma Collection of The Cancer Genome Atlas \cite{TCGA-BRCA} or the dataset of the Tumor Proliferation Assessment Challenge 2016 \cite{veta2019predicting}.

In clinical practice, \acp{WSI} are associated with private patient information such as the patient's name, age, sex, and more. This information is classified as \ac{PHI} and in most countries is protected from disclosure by government regulations such as the Health Insurance Portability and Accountability Act (HIPAA) in the United States \cite{hipaa} or the General Data Protection Regulation (GDPR) in Europe \cite{gdpr}. Hence, data anonymization is a crucial step in the process of publishing medical data \cite{willemink2020preparing,moore2015identification,bisson2023anonymization}.
% -> die nächste sich stellende Frage ist, wie sieht proper anonymization aus?
In the context of \acp{WSI}, this means removing PHI from all places where it might be present, such as the file name or any given slide tags. Also, the metadata files of the particular slide files need to be scrutinized \cite{clark2013cancer}.
% -> Possibly overlooked identity signatures
% -> With the advent of Deep Learning even formally anonymus data became identifiable
% -> X-Ray reidentificaiton

A potential issue may arise if the images contain information that can be used as a biometric identifier, such as retinal scans or fingerprints, as these can serve as 'identity signatures' that are considered PHI \cite{willemink2020preparing}.  Although fingerprints and retinal scans are commonly recognized as biometric identifiers, other clinical data such as head and neck CT scans can also contain highly patient-specific information. For example, the soft tissue kernels from head and neck CT scans can be used to reconstruct a patient's face for identification purposes \cite{willemink2020preparing}.

Given the potential of \ac{DL} algorithms to extract semantic features from images, medical image data that were not previously considered 'identity signatures' now appear to be patient-specific enough to re-identify a patient from them. For example, \cite{packhauser2022deep} demonstrated that a well-trained DL algorithm can re-identify patients from frontal chest X-ray images, even after several years have passed between the different recordings.

% -> Zusammenführen von ability to classify patterns on WSI and ability to re-identify
Given the ability of \ac{DL} models to identify complex patterns from histopathology sections and their potential for re-identifying patients from medical images, a question arises as to whether \acp{WSI} contain sufficient patient-specific information for \ac{DL} models to be used in patient re-identification.
% -> Privacy concerns as if metadata is present, one atttacker could link an anonymized image to an non-anonymized which is linked to further metadata one.
% -> Also if no metadata is available, problems might arise if the same patient is used in multiple studies with different sorts of metadata
% -> Everytime a bit more information about the patient is given wich can be pieced together in the end
The existence of such a model would raise privacy concerns, as theoretically, two ways are possible in which it could be used for disclosing \ac{PHI}: In the first scenario, if an attacker gained access to a non-anonymized slide, such a model could be used to link that image to a patient which is present in an anonymized public dataset \cite{packhauser2022deep,esmeral2022low}. If further metadata like diseases, treatment or sex is linked to the anonymized images, even more \ac{PHI} could be leaked. In the second scenario, images of the same patient are part of different datasets. This is not unlikely in the case of very rare tumors or very high tumor grades as these tend to have a lower prevalence. In this scenario a \ac{DL} model could be used to assign all these images to the same patient. If each of the previously anonymized images is linked to other metadata, merging the data could provide enough material to re-identify the patient.

Given the potential issues this work investigates whether patient re-identification is possible within large anonymized histopathology datasets using \ac{DL} algorithms. 
Our work was guided by the research questions:

\begin{itemize}
    \item[R1:] Is it generally possible to re-identify patients from histopathology whole slide images?
    \item[R2:] Is it possible to re-identify sections that have been extracted from the same patient with a considerable gap in time between the samplings?
\end{itemize}

Our contributions are the following:
\begin{itemize}
\item We design tailored experiments in an attempt to answer these research questions for the first time for the field of histopathology.
\item We identify factors contributing to the successful re-identification of anonymized histopathology samples. 
\item We formulate guidelines on how to safely publish such images, as the publication of medical data is crucial for the development of modern algorithms.
\end{itemize}

\section{Related Work}
% -> General recognition of biometric data
 The re-identification or recognition of biometric data is a long-studied field of computer vision. The re-identification of fingerprints \cite{finger1}, palms \cite{palm1,palm2}, irises \cite{iris1,iris2} or faces \cite{face1} is of high interest for modern security systems or public surveillance systems.
% -> Transition to medical data, for some cases re-identification is necessary
    % -> forensic recogniction techniques
    % -> it has also been shown, that some modalities can be used for re-identification pusposes which were never ment for that use
    % -> knee or brain MRI
    
In contrast, the identification of individuals based on medical images is primarily utilized in forensic science. Studies have demonstrated that individuals can be identified through X-ray images of their teeth \cite{teeth1,teeth2}, chest \cite{chest1,chest2}, knees \cite{knee1}, or hands \cite{hand1}. In these cases, re-identification is the intended use case. However, other studies have shown that medical data, which was never intended for re-identification purposes, like knee MRI \cite{knee2} or brain MRI \cite{brain1} scans could still be used to determine a patient's identity.
% -> Deep Learning made this even more easy, as before highly qualified skills were required to perform re-identification
    % -> Throught the use of Deep-Learning, also chest-xray and gastrointernal images became identifiable
    % -> Brief explanation of both approaches
    
With the advent of \ac{DL} methods, it became even easier to perform such re-identifications, since even relatively simple networks can extract features with high discriminative power \cite{esmeral2022low}. Two studies in this direction were done by \cite{esmeral2022low} who studied the re-identification from brain MRI and gastrointestinal endoscopic images and \cite{packhauser2022deep} who investigated the re-identification of patients from frontal chest X-ray images. In both studies, Siamese networks \cite{bromley1993signature} were employed for re-identification. The latter study approached the problem from two different directions, first as a verification problem and second as a retrieval task. For verification, the Siamese network is combined with merging layers to predict whether two X-ray images belong to the same patient. For the retrieval task, the Siamese network is trained with a contrastive loss function to extract suitable feature representations, which are then ranked based on the Euclidean distance. For both tasks, they achieved high accuracy. The verification was possible with high accuracy even with several years between the acquisition of the X-ray images.
% -> Discrimination to whole slide images, what makes that task special?
\begin{figure*}[ht]
    \centering
    \includegraphics[width=\textwidth]{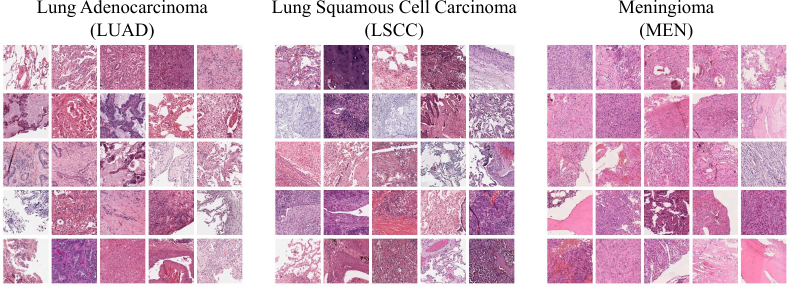}
    \caption{Overview of randomly selected patches from the three datasets used. In contrast to our in-house meningioma dataset (MEN), the lung adenocarcinoma (LUAD) and lung squamous cell carcinoma (LSCC) datasets originating from \ac{TCIA} exhibit a more pronounced visual variance. Each patch covers an area of about 0.012 square millimeters.}
    \label{fig:datasets}
\end{figure*}

Although these studies demonstrate the great potential of DL algorithms to extract patient-specific features, we believe that re-identification of histopathological sections is even more challenging for several reasons. First, the appearance of histopathology slides, even if taken from the same patient, varies substantially. Sections may differ in the amount of tumor and non-tumor tissue, the total amount of tissue contained, staining characteristics, section thickness of specimens, or by preparation artifacts such as tissue tears or folds. Secondly, the images show only a small section of tissue and not macroscopic structures such as bones or entire organs. Moreover, unlike a patient's bone structure, the morphology of successive tumors in a patient is likely to differ because tumor growth is a dynamic phenomenon that depends on the intrinsic properties of the cancer cells as well as environmental factors of the surrounding tissue microenvironment \cite{enderling2010tumor,laconi2007evolving,van2003tumor}. In addition, the sheer size of \ac{WSI} is a problem in itself, making it impossible to train standard convolutional network architectures due to limited GPU memory without using special preprocessing schemes. The most common approach to overcome this problem is to use patches extracted from the \acp{WSI} for training and inference, which implies the need for a strategy to finally aggregate the results of different patches.
% -> Related work from the field of digital pathology
    % -> mahmood lab origin of tumor
% -> mahmood lab

A related study from the field of digital pathology was published by \cite{lu2021cup} who proposed a method that predicts the origin of a tumor of unknown primary from histopathology images with high accuracy. In metastatic tumors, the primary origin is sometimes unclear but plays an important role in tumor treatment. They used an attention-based \ac{MIL} pipeline, which uses precomputed features similar to \cite{lu2021clam}.
% -> transition, end of related work
In contrast to tracing a tumor back to its origin, we want to trace a histological slide back to a tumor or a patient. This leads to a larger search space and a more complex problem. To solve this problem, we investigate two different strategies.

\section{Materials and Methods}

\subsection{Datasets}
In this study, we utilized three distinct datasets of which two are publicly available. Those two datasets, namely \ac{LUAD} \cite{LUAD} and \ac{LSCC} \cite{LSCC}, were obtained from \ac{TCIA} \cite{clark2013cancer}. In the remainder of this paper, these datasets will be referred to as the \ac{LUAD} dataset and the \ac{LSCC} dataset. These datasets were scanned at a resolution of 0.5 microns per pixel and were obtained from various pathology centers. We restricted our analysis to slides of patients for which at least two slides were available, resulting in 1059 images of 226 patients for the \ac{LUAD} dataset and 1071 images of 209 patients of the LSCC dataset.
% -> Men dataset

The third dataset is an in-house dataset of slides of \ac{MEN} tissue, collected as consecutive selection from the diagnostic archive of the Department of Neuropathology, University Hospital Erlangen, Germany. We received ethics approval from the institutional review board of the Medical Faculty of FAU Erlangen-Nürnberg (AZ 92\_14B, AZ 193\_18B). All slides stem from patients where a surgical resection of the tumor was performed. After resection, the tissue was fixated, then embedded in paraffin and finally stained with standard hematoxylin and eosin stain using an autostainer. The slides were scanned with a resolution of 0.22 microns per pixel, using a Hamamatsu NanoZoomer S60. As with the \ac{TCIA} datasets, we only included patients for which at least two slides were available, resulting in 979 slides from 244 patients. This dataset will be referred to as the MEN dataset in the following.
% -> heterogeneity of multicenter datasets

Figure~\ref{fig:datasets} compares patches from the datasets used in our study. In contrast to our in-house dataset where all slides originate from a single center, the multicenter, public datasets have a more heterogeneous appearance.

% -> why use in house dataset
% -> In this subsection, we can cite "The impact of site-specific digital histology signatures on deep learning model accuracy and bias"
    % -> The sites where the slides are from can be identified from the images
    % -> This shows the need for an single center data sets, as here such cues are not present
The need for the use of our in-house dataset arises from research question R2, for which a dataset with information about the time of resections is required. This information was not available for the \ac{LUAD} and \ac{LSCC} datasets. Of the 244 patients in the \ac{MEN} dataset, 39 had undergone multiple resections. Of those, 74\,\% underwent two resections, 8\,\% underwent three resections and  18\,\% underwent four or more resections. Although the exact dates of the resections were not available, we estimated the time between them from the running numbering of cases per year. The resulting average duration between surgical resections was 1,380 days with a standard deviation of 1,336 days. The shortest interval between two consecutive resections was 30 days, while the longest was 4,505 days.

\begin{figure}[ht]
    \centering
    \includegraphics[width = \linewidth]{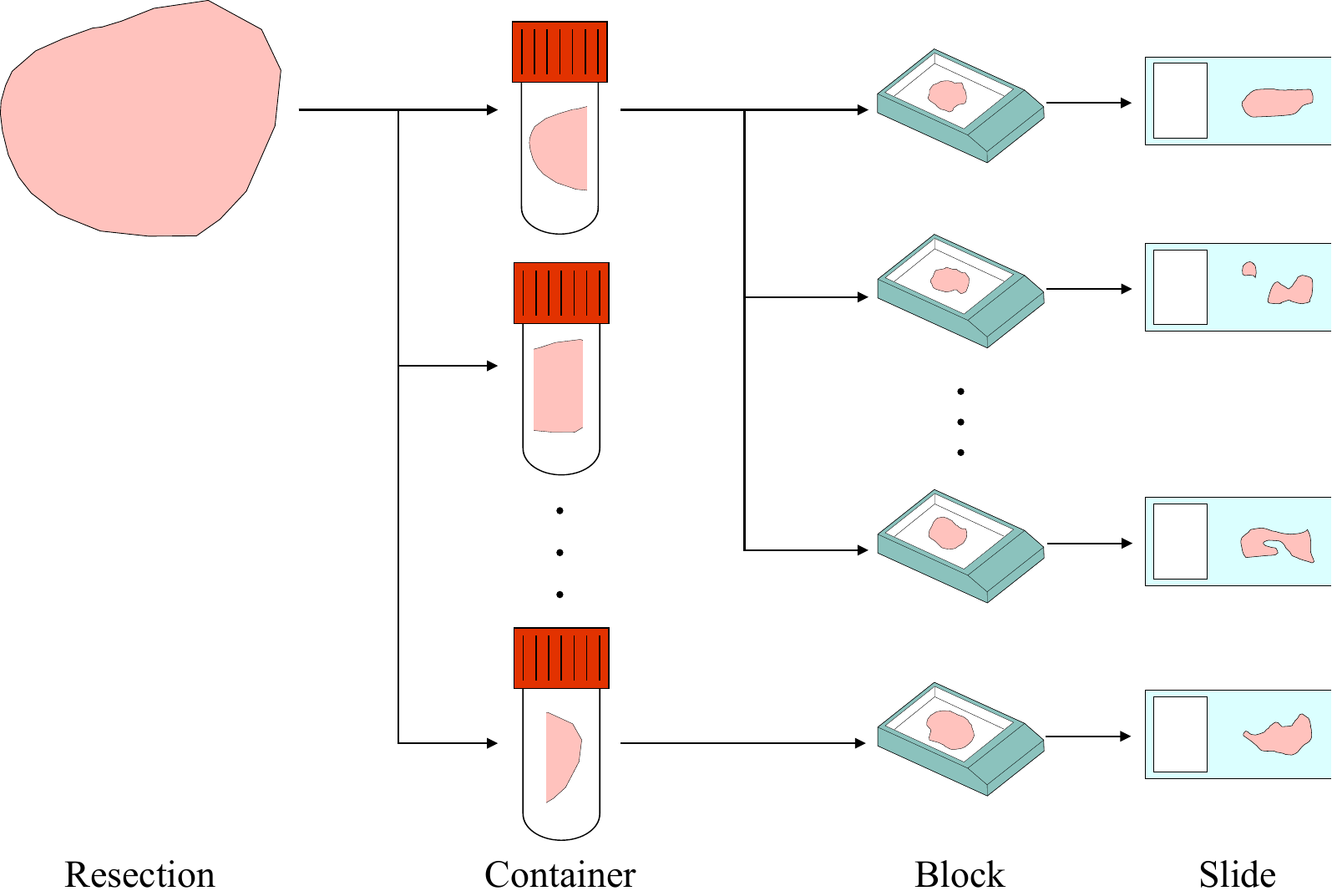}
    \caption{Scheme of the tissue preparation procedure used to prepare the slides in the in-house meningioma (\ac{MEN}) dataset. A resection can be divided into one or more containers, each of which can be further divided into one or more blocks.  However, only one slide from each block is included in the data set.}
    \label{fig:scheme_histo}
\end{figure}

% -> Tissue preperation
In the \ac{MEN} dataset, a single resection can be associated with multiple images. This is due to the preparation process shown in Figure~\ref{fig:scheme_histo}. At the beginning of the preparation process, a resection may be cut into several pieces that are placed in different containers for fixation. After fixation, the tissue of each piece is embedded in a single paraffin block. Depending on its size, a sample may result in multiple blocks. From each block, one histology slide is prepared. Due to the size of the resection, the content of one container can also result in only one block as depicted in Figure~\ref{fig:scheme_histo}. All images that are associated with the same resection are from different blocks, i.e., no consecutive sections from the same block were used in the \ac{MEN} dataset.
% Tabelle mit Infos

\subsection{Methods}
% -> Motivation, why classification? 
The re-identification problem is formulated as a subject classification task, where each patient represents one class. In a hypothetical attack scenario, the training data would consist of an anonymized public dataset, while the test data would be slides for which the patient of origin is known.

% -> Why not verification
Another possible approach could be to formulate the problem as a two-class classification problem, where we classify if two given slides belong to the same patient or not \cite{packhauser2022deep}. However, we consider the subject classification task to be a more straightforward approach. If one wanted to re-identify a slide by verification, one would have to query every slide in the training data. In contrast, our approach requires only a single query but does not apply to targets not included in the anonymized dataset.
% -> Begin description of methods 

In this study, we compare two of the most common classification architectures to solve the classification problem at hand~\footnote{Code will be made available online on github upon acceptance.}. Since \acp{WSI} are typically too large to fit into GPU memory at once, all algorithms were trained on patches extracted from \acp{WSI}.
% Patch Classifier

\subsubsection{Model Architectures}
The first approach comprised a classification network with a ResNet18 as the backbone and a fully connected layer with $N$ output neurons, where $N$ is equal to the number of patients. Each patch is classified separately as belonging to one of $N$ patients during training. In the inference, all patches of one slide were classified, and a final classification result was aggregated using the argmax function. Since the appearance of patches extracted from one slide can be highly heterogeneous, we do not assume that all patches contain semantically meaningful information. Therefore, we consider the training of this classifier as training with noisy labels.
% MIL Classifier
\begin{figure}[ht]
    \centering
    \includegraphics[width=\linewidth]{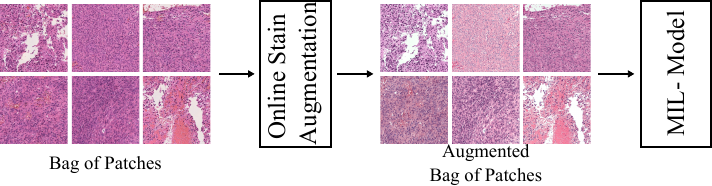}
    \caption{Scheme of how the online stain augmentation was applied during \ac{MIL} training. During training, each of the images within one bag was augmented separately.}
    \label{fig:mil_training}
\end{figure}

For the second approach, we used a \ac{MIL} algorithm. This is a variation of weakly supervised learning where instead of labeling each instance, a bag of instances is given a single label. It is assumed that within a bag, the bag label is present, but that not all instances carry the label \cite{SelfSupervisedHisto}. We used a gated attention network as \ac{MIL} pooling function and a set of parallel classifiers for the final prediction as described in \cite{lu2021clam}. Contrary to Lu et al., we did not use precomputed features to train the attention network, as this would have made it impossible to use online augmentation, which is important to mitigate overfitting. Instead, we trained the attention- and classification network on randomly selected bags of a size of 40 patches, where each patch is augmented separately as depicted in Figure~\ref{fig:mil_training}. This online augmentation enabled us to train the attention network with a higher variance than would have been possible with a set of precomputed features.

 % Backbone and pretraining
Similar to \cite{lu2021clam}, our \ac{MIL} network also incorporates a frozen ResNet18 as the backbone. We examined three different feature encoders for this network. The first encoder is the stem of the network trained with the patch classification model. The second is a backbone trained using \ac{SSL} on a large corpus of histopathology images from different organs \cite{SelfSupervisedHisto}. For last the feature encoder, we used a network pre-trained on ImageNet.

\subsubsection{Preprocessing}
\begin{figure}[ht]
    \centering
    \includegraphics[width=8cm]{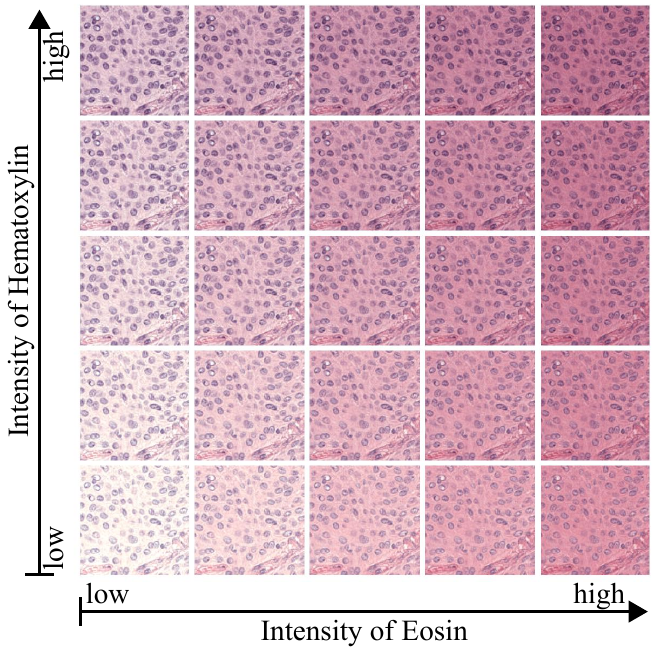}
    \caption{Given are versions of the same patch to which different intensities of stain augmentation were applied. A stain augmentation based on the Macenkos stain normalization method was used. The non augmented patch is given in the center of the grid.}
    \label{fig:marcenko}
\end{figure}
% Different Aggregations
%During training and inference, only patches which had sufficient tissue coverage were used.
% Stain Augmentation
% -> Further ellaborate on that!!!!
% -> Bilder welche an einem Tag erstellt wurden sehen ähnlicher aus
  
Since the tissue perpetration procedure is known to leave center-specific traces on the slides which influence the generalization of the trained models \cite{keller2023tissue}, we applied dedicated augmentation strategies to counteract these influences. To prevent the network from using cues originating from tissue preparation procedures or differences in color representation by the used scanner, we used stain augmentation during training based on the method for stain normalization proposed by \cite{macenko2009method}, which is based on the Beer-Lambert law. First, the absorbance or \ac{OD} is calculated from the normalized RGB color vector $I$ as follows:
\begin{align*}
    \label{eq1}
    \mathrm{OD} &= -\log_{10}\left(I\right) \\
\end{align*}
From OD, the stain and saturation vectors $S$ and $V$ are then calculated using color deconvolution. Augmentation is performed by perturbing the saturation vector $S$ as similar to ~\cite{otalora2022stainlib}:
\begin{gather*}
    S' \leftarrow \alpha S + \beta
\end{gather*}
Where $S'$ is the augmented saturation vector, and $\alpha$ and $\beta$ are drawn from a uniform distribution between $1-\lambda\leq\alpha\leq1+\lambda$ and $\lambda\leq\beta\leq\lambda$. For all our experiments we used $\lambda = 0.2$. Afterward, the augmented image is reconstructed using the augmented saturation vector $S'$. This augmentation was applied to every patch used during training. During the training of the \ac{MIL} models, each image within one bag was augmented separately. Examples of augmented patches are shown in Figure~\ref{fig:marcenko}.

% Other Augmentations
In addition, we also included horizontal and vertical flipping in our augmentation pipeline.
% Magnification level
For all experiments, we used patches with a resolution of 0.88 \ac{mpp}, as we found this to be the optimal resolution for the task at hand in pre-study experiments. The detailed results of these experiments can be found in A.
% Sampling
\subsubsection{Sampling and Training Parameters}
For higher computational efficiency, we used Otsu's adaptive thresholding \cite{otsu1979threshold} method on a size-reduced grayscale version of the \ac{WSI} to generate a segmentation map for tissue. During training, only patches with a tissue coverage above 70\,\% were used.
% Training hyperparameters
All models in this study were trained until convergence was observed on the validation dataset. For all experiments, we used the Adam optimizer with a learning rate of $10^{-4}$ and a cross-entropy loss. The best model was then selected retrospectively based on the validation loss.

\section{Experiments}
\begin{figure*}[ht]
    \centering
    \includegraphics[width=\textwidth]{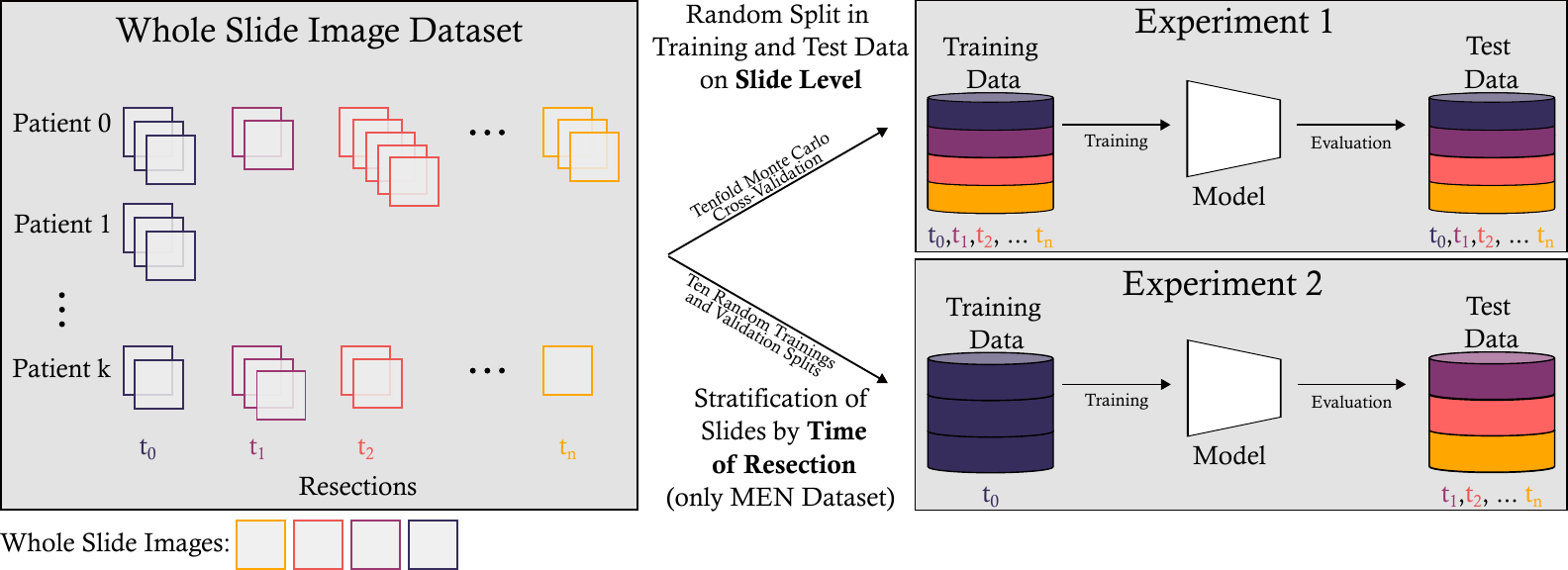}
    \caption{Overview of the experimental setup of Experiments 1 and 2. Experiment 1 involved a tenfold Monte Carlo cross-validation. In Experiment 2, the slides from the earliest resection were used for training, while all images from later resections were used in a hold-out test dataset. To increase the statistical validity of the results of Experiment 2, ten models for each algorithm were trained on ten randomly selected training and validation splits drawn from the earliest resection of each patient.}
    \label{fig:experiments}
\end{figure*}

Several experiments were conducted to address the research questions formulated in the beginning. The experimental setup of Experiments 1 and 2 is depicted in Figure~\ref{fig:experiments}. 

\subsection{Experiment 1}
Experiment 1 was designed to address research question R1. To investigate whether re-identification of tumors based on histopathology images is possible at all, we randomly divided the slides of each dataset into training, validation, and test sets as shown in Figure~\ref{fig:experiments}. This corresponds to the case where nothing is known about when a tumor was resected. To increase the statistical power of our experiments, we performed a tenfold Monte Carlo cross-validation. This experiment was performed on all three datasets,  \ac{LUAD}, \ac{LSCC}, and \ac{MEN}. We examined the performance of both the patch-based approach and the MIL approach. 

For the MIL approach, we also investigated the performance using different models used for the frozen decoder. This investigation was solely conducted on the \ac{MEN} dataset because of its homogeneous appearance compared to the \ac{LUAD} and \ac{LSCC} datasets. We assume that strong cues related to preparation and staining are less pronounced in the \ac{MEN} dataset. Consequently, this leads to a higher interpretability of the results.

\subsection{Experiment 2}
To answer research question R2, we stratified the slides from the \ac{MEN} dataset by the time of resection. This way, we aim to determine if re-identification is still possible even if the slides of the training and test sets stem from different resections. We did this by training the models on all slides belonging to a patient's earliest resection (see Figure~\ref{fig:experiments}). For testing, all slides of later surgical resections were used. We trained our models ten times on different training and validation splits drawn from the slides belonging to the earliest resection of every patient. Patients without images from a later resection were still included in the training set. Consequently, the algorithms still need to discriminate between all patients present in the training and validation data, as in Experiment 1. For these patients, there are no slides in the hold-out test set. Since the \ac{LUAD} and \ac{LSCC} datasets do not contain any information about when the resections were performed, we only evaluated this experiment on the \ac{MEN} dataset.
\subsection{Post hoc Analysis of Experiments 1 and 2}

We conducted an additional analysis of our models to identify the factors that influence the model's predictions. The \ac{MEN} dataset was chosen for analysis due to its homogeneity, in comparison to the \ac{LUAD} and \ac{LSCC} datasets.

We examined the L2 distances between the feature embeddings of the test samples and the respective training samples. We calculated the average embedding for each patient from all training images of that patient. From here on, we will refer to this mean embedding as the latent space anchor. We calculated the distance from the embedding of each sample in the test set to its respective latent space anchor. For the computation of the embeddings, we used the feature encoders that were part of the MEN-Patch and MEN-MIL models.
\section{Results}
% -> Introduction Resutls, Metrics
For all experiments, we report the recall@1, recall@5, the precision and the $F_1$ score. These values are always the average values over all classes. When considering recall@n, it means that for an algorithm's predictions to be considered correct, the searched patient has to be included among the $n$ patients with the highest-ranked predictions based on the classification score. In a multiclass classification problem, the average recall@1 equals the balanced accuracy. For comparison, the probability of selecting the right patient by chance when assessing $N$ patients is also given for each dataset.
\subsection{Results of Experiment 1}
% -> accuracys of patch methods on different datasets
On all three datasets, the methods demonstrated satisfactory performance in re-identifying the patients based on histology slides. Detailed results can be found in Table \ref{tab:R1}. On the \ac{MEN} dataset, the patch-based MEN-Patch model achieved the highest recall@1 with 65.25 \%. The \ac{MIL} model, MEN-MIL, performed slightly worse, with a recall@1 of 61.13 \%. For the recall@5, the values were 80.59 \% and 77.53 \%, respectively.
% -> pretraining strategies MIL
When comparing different feature encoders used as the feature extractor of the \ac{MIL} approach, the MEN-MIL model, which includes the feature stem trained as part of the MEN-Patch model, yielded the most favorable results. The MIL model, which includes a feature encoder trained on ImageNet (MEN-MILImageNet), achieved a recall@1 of 30.63 \%. The MIL model that comprises a feature stem trained through SSL (MEN-MILSSL) achieved a recall@1 of 24.85 \%.
% -> Results on LUAD & LSCC datasets
For the two public datasets, we found very similar performances for the patch-based and \ac{MIL} approaches. On the \ac{LUAD} dataset, the patch-based and the \ac{MIL} approach achieved a recall@1 of 53.61 \% and 53.18 \%. On the \ac{LSCC} dataset, we found a recall@1 of 55.06 \% and 56.31 \%, respectively. In contrast to the recall@1, which was slightly higher on the \ac{MEN} dataset, all models achieved very similar results for the recall@5.

\begin{table}[ht]
    \centering
    \caption{Results of Experiment 1. The respective means and standard deviations of the tenfold Monte Carlo cross-validation are given. In a multiclass classification problem, the mean recall is equal to the balanced accuracy. Random probability is the probability of selecting the correct patient by random guessing.}
    \resizebox{\linewidth}{!}{
    \begin{tabular}{lcccccccc}
    \hline
    Method & recall@1 & recall@5 &  precision &  $F_1$ score \\
    \hline
    MEN-Patch & $65.25\% \pm 6.49\%$ & $80.59\% \pm 4.02\%$ & $63.79\% \pm 6.30\%$ & $62.31\% \pm 6.45\%$ \\
    MEN-MIL & $61.13\% \pm 7.00\%$ & $77.53\% \pm 3.80\%$ & $58.41\% \pm 7.71\%$ & $57.49\% \pm 7.54\%$ \\
    MEN-MILImageNet & $30.63\% \pm 4.37\%$ & $52.12\% \pm 7.05\%$ & $28.07\% \pm 5.06\%$ & $27.10\% \pm 4.64\%$ \\
    MEN-MILSSL & $24.85\% \pm 2.38\%$ & $48.19\% \pm 2.13\%$ & $20.99\% \pm 1.98\%$ & $20.80\% \pm 1.98\%$ \\
    Random probability &$0.40\%$&$9.58\%$& & \\
    \hline
    LUAD-Patch & $53.61\% \pm 5.73\%$ & $76.20\% \pm 4.86\%$ & $52.39\% \pm 5.92\%$ & $50.30\% \pm 6.07\%$ \\
    LUAD-MIL & $53.18\% \pm 6.59\%$ & $76.09\% \pm 4.66\%$ & $52.74\% \pm 5.78\%$ & $50.16\% \pm 6.43\%$ \\
    Random Probability &$0.48\%$& $11.06\%$& & \\
    \hline
    LSCC-Patch & $55.06\% \pm 3.33\%$ & $80.34\% \pm 2.93\%$ & $53.04\% \pm 3.19\%$ & $51.16\% \pm 3.28\%$ \\
    LSCC-MIL & $56.31\% \pm 3.54\%$ & $79.43\% \pm 2.66\%$ & $54.01\% \pm 3.91\%$ & $52.30\% \pm 3.66\%$ \\
    Random Probability &$0.42\%$ & $10.29\%$ & & \\
    \hline
    \end{tabular}}
    \label{tab:R1}
\end{table}

\subsection{Results of Experiment 2}
When the models were trained on the earliest resection and tested on later resections of the \ac{MEN} dataset, the performance dropped remarkably compared to Experiment 1 (see Table \ref{tab:R2}). The highest performance among the compared methods was observed for the \ac{MEN}-MIL method, with a recall@1 of 13.53\,\% and a recall@5 of 22.13\,\%. In contrast to the first experiment, the \ac{MIL}-based approaches exhibited better performance compared to the patch-based model. Similar to Experiment 1, the MEN-MILSSL model yielded the lowest results. Even though the individual results were lower than in Experiment 1, they all remained considerably above the respective probabilities of random guessing.

\begin{table}[ht]
\centering
\caption{Results of Experiment 2. In a multiclass classification problem, the balanced accuracy equals the average recall. Random probability is the probability of selecting the correct patient by random guessing.}
\resizebox{\linewidth}{!}{
\begin{tabular}{lcccccccc}
\hline
Method & recall@1 & recall@5 & precision & $F_1$ score \\
\hline
MEN-Patch & $9.72\% \pm 2.17\%$ & $18.41\% \pm 1.93\%$ & $13.27\% \pm 3.41\%$ & $10.08\% \pm 2.38\%$ \\
MEN-MIL & $13.53\% \pm 6.59\%$ & $22.13\% \pm 10.40\%$ & $17.04\% \pm 6.87\%$ & $14.30\% \pm 6.63\%$ \\
MEN-MILImageNet & $10.31\% \pm 1.87\%$ & $20.88\% \pm 1.43\%$ & $12.30\% \pm 1.84\%$ & $9.35\% \pm 1.24\%$ \\
MEN-MILSSL & $6.42\% \pm 2.20\%$ & $16.66\% \pm 2.45\%$ & $6.11\% \pm 2.01\%$ & $5.00\% \pm 1.74\%$ \\
Random Probability & $0.40\%$ & $9.58\%$ &  &  \\
\hline
\end{tabular}}
\label{tab:R2}
\end{table}

\subsection{Results of the Post hoc Analysis of Experiments 1 and 2}
\begin{figure}[ht]
    \centering
    \includegraphics[width=\linewidth]{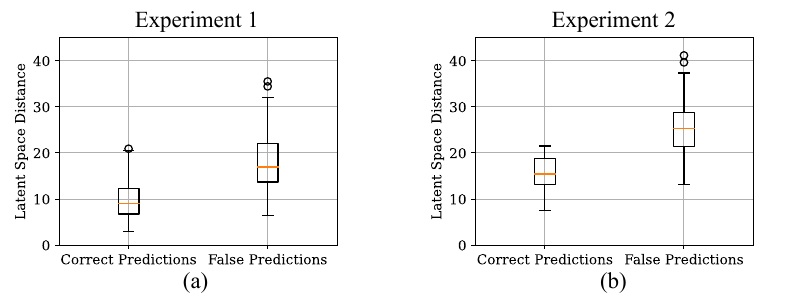}
    \caption{Distances between test samples and their respective latent space anchors. Sub-figure a) shows the distances for Experiment 1 and sub-figure b) shows the distances for Experiment 2. In general, correctly classified samples are closer to their respective latent space anchors.}
    \label{fig:R3_1}
\end{figure}

 Figure~\ref{fig:R3_1} illustrates the L2-distance between latent space embeddings of test samples and their corresponding latent space anchors for Experiments 1 and 2. In both experiments, the correctly classified samples were observed to have a considerably smaller distance to their respective latent space anchors than the misclassified samples. Additionally, the distances between test samples and their latent space anchors were considerably smaller in Experiment 1 than in Experiment 2.
 % In Experiment 1, the models were trained on the earliest resection of each patient and tested on the slides of subsequent resections. This was observed for both correctly and incorrectly classified samples.
 
\section{Discussion}
% -> generell beantworten das es funktioniert 

    % -> suprisingly high accuracy
    % -> as indicated by the results of the first exp
% -> However exp 2 indicates that tumor reidentiication is possible but not patient re-identification
% EXP1
% 1. Hat bei allen funktioniert
% 2. Patch == MIL
% 3. On Tasnk besser als ImageNet und SSL -> ImageNet aber immernoch gut
% -> Stain wenig Einfluss wegen Stainaugmenatation
    % -> MEN besser als LUAD und LSCC
The results of Experiment 1 show that re-identification of tumors from histopathology images is generally possible. Yet, in this experiment, there might be a huge number of covariant factors that acted as supplementary cues for the model and could explain this surprisingly high performance. In particular, even though none of the slides were direct and consecutive sections, the majority will have been resected from the same tumor, prepared on the same day, by the same lab. Since the HE dye changes its properties throughout processing many samples with it, this could also be incorporated into the visual properties of the slide, leaving an identifiable, visual time stamp of some kind on the slides. Although we attempted to mitigate the influence of these factors through the use of stain augmentations, there may still be residual traces that the model could use to re-identify the slides \cite{keller2023tissue}. Even more covariant factors might be present in multicentric datasets like the \ac{LUAD} and \ac{LSCC} datasets as different labs use different stains from different manufacturers, dissolved in water with different chemical attributes. Centers may also differ in the tools used to prepare slides, such as microtomes for cutting slides from tissue blocks, or whether an autostainer or manual staining is used to stain slides \cite{Howard2021,dehkharghanian2023biased}. Even the tissue fixation procedure used pre-analytically can influence tissue quality and hence leave traces that could in some cases be attributed to a certain clinical environment.
 % This is a privacy problem of itself. If the preperation procedure really leaves this amount of information on the slides, perhaps steps have to be taken into the direction of clearing them before publication.
%Although we believe that the stain augmentation method we used prevented the models from distinguishing patients based on mere differences in staining, the sections might share other properties, such as the thickness of the sections which has an influence overall brightness of the resulting WSI. Therefore, the simpler patch method was able to learn sufficient features to differentiate between patients.\\
% -> The lower performance on the LUAD and LSCC Datasetes as a argument for the well functioning of the stain Augmentation
Since the \ac{MEN} dataset was processed in a single center, the covariant factors should not have had much influence on the results on this dataset. Therefore, we originally expected that re-identification would be more successful in the \ac{LUAD} and \ac{LSCC} datasets, as the models might also use the center-specific traces for re-identification. However, we found a lower performance for these datasets, indicating that the models did not heavily rely on center-specific cues to fulfill the task. This potentially implies that the stain augmentation might have been a successful tool to prevent the model from using stain differences for re-identification. 
%However, this might also be caused by the models over-relying on center-specific cues in the \ac{LUAD} and \ac{LSCC} case. In the single-specific case the classification error might distribute more evenly across other classes than in the multi-centric case, where a clustering might be expected. Unfortunately, since this information is not available for neither dataset, we can not test this hypothesis.

If the models focused more on morphological features to do the re-identification, the high heterogeneity of image qualities in public repositories \cite{janowczyk2019histoqc} like the \ac{LUAD} and \ac{LSCC} datasets could have even led to lower performances in contrast to the \ac{MEN} dataset. 

Another fact that points toward the importance of morphological features are the results of our preliminary investigation in A, where we searched the optimal magnification level on the \ac{MEN} datasets. If the models had only relied on differences in staining, one could assume that the performance differences between the different magnification levels would have been less pronounced. However, our results showed that a medium resolution (0.88 mpp) is best suited to the task. At this resolution, both local features and the general structural composition of the tumor are accessible to the model.
% --> Scaling ansprechen: Es liegt nicht daran, dass ein Datensatz deutlich mehr Bilder pro Patient hat.
% 

% -> Patch und Men ähnliche Performance, warum?
%Our results showed similar performances for the patch-based and \ac{MIL}-classifiers on the different datasets. This might be because the attention network of the \ac{MIL}-classifier learned a very similar aggregation function to the argmax function that is part of the inference of the patch-classifier.
We found similar performance for the patch-based and \ac{MIL}-classifier on the different datasets, with still a consistent edge of the approach that aggregates the patch classifications over the \ac{MIL} approach. This might be a consequence of the argmax aggregation, which has a receptive field of the complete image and can compare the individual votes for all patches. In contrast, the attention mechanism, which is used for aggregation in the \ac{MIL} approach, only calculates an instance-based weight for each patch and has thus a very limited receptive field of only one instance.

% -> Analysis of pretraining strategies
The analysis of various feature encoders for the \ac{MIL} model showed that the models trained specifically for the given task consistently outperform the other models. However, both the \ac{SSL}-trained model and the model that was solely trained on ImageNet demonstrated notable accuracy in discriminating patients. The comparison of the ImageNet and \ac{SSL} trained models indicates that the features learned by the \ac{SSL} model were less discriminative for our downstream task than the ImageNet features. In \ac{SSL}, the learning process is guided by the pseudo-labels generated from the data itself. If they do not reliably capture discriminative features relevant to the downstream task, this may lead to less effective learning compared to a model trained on ImageNet, which is a large-scale dataset with manually annotated, high-quality labels.

% Hier noch nach einer Referenz suchen und dann sowas wie "This is in line with....."
% EXP2
% -> drop in performance because cues discussed above were missing
% -> but still some tumors identifiable -> incomplete resections
Experiment 2 aimed to assess the feasibility of re-identifying sections extracted from the same patient at different time points. The results of this experiment demonstrated considerably lower performance compared to Experiment 1. One possible explanation for this difference in results is that the slides were prepared at different time points. Therefore, unlike in Experiment 1, the resections in Experiment 2 showed covariant factors to a much lesser extent. Despite this, the classifier was still able to assign some tumors to the correct patient with a higher accuracy than if the slides had been assigned randomly. This could be due to an incomplete resection of a primary tumor, resulting in morphological similarities that the models used for re-identification. Another reason could be that the latter tumor is a metastasis of the primary one, and they are thus sharing similar morphological patterns.

% EXP3
% -> zuerst auf Abbildung R3_1 eingehen.
    % -> sinnvoller Latent Space
    % -> Ex2 höhere Distanz weil andere Container
The post hoc analysis of Experiments 1 and 2 gave insights into the latent space organization that the model used for the successful re-identification. Figure~\ref{fig:R3_1} shows that in Experiments 1 and 2, correctly classified samples had smaller latent space distances to their latent space anchors compared to misclassified samples. This indicates that the models were able to encode features into a semantically meaningful latent space, as correctly classified samples share more visual features, resulting in a smaller latent space distance. In Experiment 2, the distances in the latent space were significantly higher for both correctly and incorrectly classified samples compared to Experiment 1. This could be due to the absence of covariant visual features related to slide preparation between the training and test images in Experiment 2, as previously described. Since the correctness of the classification is strongly correlated with the latent space distance, distance-dependent re-identification approaches as described in \cite{packhauser2022deep,esmeral2022low} might also have been feasible for solving the task at hand.

One limitation of our work is the limited number of patients for which we had data from different resections. This limits the implications of our work regarding the question if re-identification is possible between different resections of the same patient. Another limitation of our work is that since the \acp{WSI} contained mostly tumor, we can't make implications about the re-identification of healthy tissue. Moreover, we only investigated whole slide images originating from two different organs (brain, and lung). Given the strong consistency of our results on the three data sets, we are confident that the results generalize to other tumor types, however. 
%#######
% Nochmal darauf eingehen das wir unterschiedliche Blöcke haben
%#######

\section{Recommendations}
\begin{figure*}[ht]
    \centering
    \includegraphics[width=\textwidth]{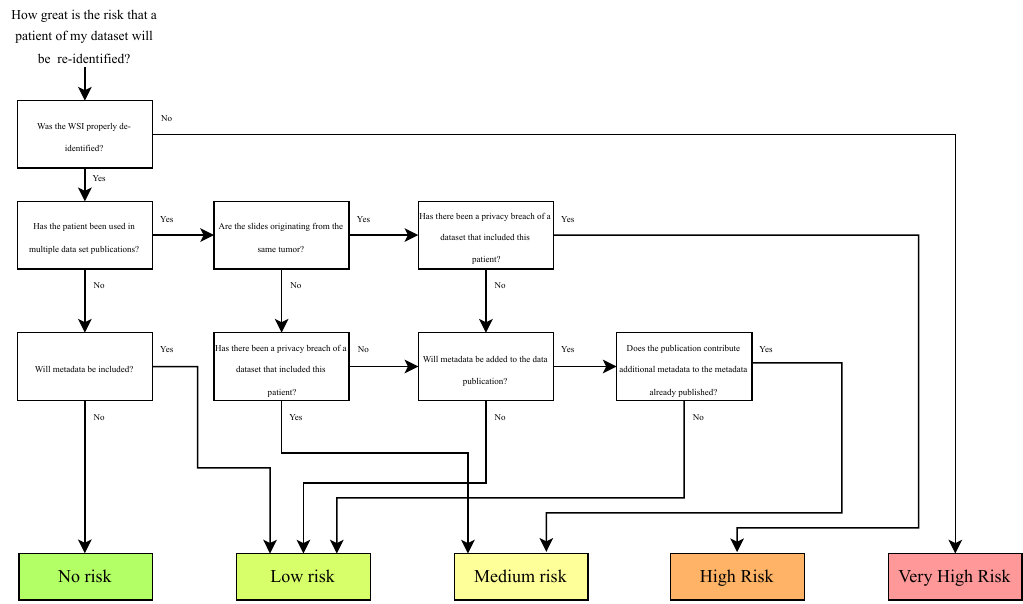}
    \caption{Risk assessment scheme for estimating patient privacy risks when publishing histopathology images.}
    \label{fig:risk_assessment}
\end{figure*}

Sharing histopathology images is crucial for the development of \ac{DL} algorithms for digital pathology, despite possible concerns about patient privacy. Therefore, we propose a risk assessment scheme (shown in Figure~\ref{fig:risk_assessment}) based on our findings to evaluate the risk to patient privacy before publishing the data.

The first prerequisite for safely sharing histopathology slides is to properly de-identify the images from any present \ac{PHI} as it is practiced when publishing data to public image repositories like the TCIA \cite{clark2013cancer}. This includes the removal of \ac{PHI} from the metadata of the slide file itself, clearance of any patient identifiers from the filenames, removal of any scanned slide tags from the image files (from the thumbnail image or the border of the region selected for scanning), and checking whether no other \ac{PHI} are directly part of the image itself. If those steps are not carefully met, one risks the violation of GDPR or HIPAA guidelines and puts the patient's privacy at considerable risk.

Additionally, it is important to determine if data from the patient has been included in previous dataset publications. As stated earlier, this is not unlikely in the case of images of rare diseases or images of tumors of a high grade because of their low prevalence. If the images have not been previously published and the data has been properly anonymized, there is no direct risk to the patient's privacy. 

The deterioration of the results in Experiment 2 compared to Experiment 1 shows that re-identification is significantly more successful when performed between images of the same resection, even if non-consecutive slides are involved. Therefore the risk of a possible re-identification depends considerably on whether the published images originate from the same tumor. 
The risk of re-identification is reduced, if sections from another tumor of the same patient are used, should these be available.

Privacy breaches of datasets in which a patient was included also have to be taken into account. A hypothetical attacker might gain access to the information and link it to some currently anonymized data, which would result in the disclosure of even more private data \cite{packhauser2022deep,esmeral2022low}.

Our results demonstrate that sections from the same tumor can be re-identified with non-negligible performance. As a consequence, if medical metadata is to be published along with the images, it is important to keep track of which images and metadata of which patient have already been published. If the same patient is used in different publications with images from the same tumor but with different metadata, the tumor tissue might be used as a key to link the datasets and hence the risk of re-identification increases \cite{sweeney2002k}. The more datasets with metadata can be combined using the tumor characteristics serving as a unified key, the higher the likelihood of re-identification.

\section{Conclusion}

This work demonstrates that re-identification of patients from histopathology images of resected tumor specimens is possible, with some limitations. As long as the slides originate from the same tumor, we can re-identify the patients with considerable accuracy (as can be seen in Table \ref{tab:R1}). If the slides were resected at different points in time, the accuracy is considerably lower (see Table \ref{tab:R2}). 
A successful resection should completely remove the tumor, and hence a later resection resembles a regrowth of an incomplete resection or a new tumor of potentially different pathogenesis and mutational pattern. Our results indicate that the strong performance drop could be linked to different morphological tumor characteristics. Consequently, our approach is more likely to identify tumors than patients. 

Which visual factors in particular contribute to the re-identification is a question for future research. However, even if the models would heavily rely on traces related to slide preparation to re-identify the slides, this would threaten patient privacy. Therefore if these factors would imprint some kind of implicit visual time stamp, future work can focus on how to remove these traces from the slides.

Our results indicate that the safest way of publishing histopathology images is to only use each patient in one data publication, as tracing across datasets and hence recombination of multiple meta and image datasets is feasible, especially if slides originating from the same tumor are used in different datasets.   

% \section{Author Contributions}
% Jonathan Ganz: Conceptualization, Methodology, Software, Investigation, Writing - Original Draft, Visualization. Jonas Ammeling: Writing - Review \& Editing, Conceptualization. Samir Jabari: Resources, Data Curation, Supervision. Katharina Breininger: Writing - Review \& Editing, Conceptualization, Supervision. Marc Aubreville: Writing - Review \& Editing, Conceptualization, Supervision.

%\bibliographystyle{elsarticle-num}
\bibliographystyle{splncs04}
\bibliography{Reidentification}

\begin{thebibliography}{10}
\providecommand{\url}[1]{\texttt{#1}}
\providecommand{\urlprefix}{URL }
\providecommand{\doi}[1]{https://doi.org/#1}

\bibitem{aubreville2023mitosis}
Aubreville, M., Stathonikos, N., Bertram, C.A., Klopfleisch, R., Ter~Hoeve, N., Ciompi, F., Wilm, F., Marzahl, C., Donovan, T.A., Maier, A., et~al.: Mitosis domain generalization in histopathology images—the midog challenge. Medical Image Analysis  \textbf{84},  102699 (2023)

\bibitem{bisson2023anonymization}
Bisson, T., Franz, M., Dogan~O, I., Romberg, D., Jansen, C., Hufnagl, P., Zerbe, N.: Anonymization of whole slide images in histopathology for research and education. Digital Health  \textbf{9},  20552076231171475 (2023)

\bibitem{bromley1993signature}
Bromley, J., Guyon, I., LeCun, Y., S{\"a}ckinger, E., Shah, R.: Signature verification using a" siamese" time delay neural network. Advances in neural information processing systems  \textbf{6} (1993)

\bibitem{hipaa}
{Centers for Disease Control and Prevention}: Health insurance portability and accountability act of 1996 (hipaa) (2023), \url{https://www.cdc.gov/phlp/publications/topic/hipaa.html}, accessed: 08 Aug 2023

\bibitem{chest2}
CHO, H., ZIN, T.T., SHINKAWA, N., NISHII, R.: Post-mortem human identification using chest x-ray and ct scan images. International Journal of Biomedical Soft Computing and Human Sciences: the official journal of the Biomedical Fuzzy Systems Association  \textbf{23}(2),  51--57 (2018)

\bibitem{SelfSupervisedHisto}
Ciga, O., Xu, T., Martel, A.L.: Self supervised contrastive learning for digital histopathology. Machine Learning with Applications  \textbf{7},  100198 (2022)

\bibitem{clark2013cancer}
Clark, K., Vendt, B., Smith, K., Freymann, J., Kirby, J., Koppel, P., Moore, S., Phillips, S., Maffitt, D., Pringle, M., et~al.: The cancer imaging archive (tcia): maintaining and operating a public information repository. Journal of digital imaging  \textbf{26},  1045--1057 (2013)

\bibitem{coudray2018classification}
Coudray, N., Ocampo, P.S., Sakellaropoulos, T., Narula, N., Snuderl, M., Feny{\"o}, D., Moreira, A.L., Razavian, N., Tsirigos, A.: Classification and mutation prediction from non--small cell lung cancer histopathology images using deep learning. Nature medicine  \textbf{24}(10),  1559--1567 (2018)

\bibitem{dehkharghanian2023biased}
Dehkharghanian, T., Bidgoli, A.A., Riasatian, A., Mazaheri, P., Campbell, C.J., Pantanowitz, L., Tizhoosh, H., Rahnamayan, S.: Biased data, biased ai: deep networks predict the acquisition site of tcga images. Diagnostic pathology  \textbf{18}(1),  1--12 (2023)

\bibitem{enderling2010tumor}
Enderling, H., Hlatky, L., Hahnfeldt, P.: Tumor morphological evolution: directed migration and gain and loss of the self-metastatic phenotype. Biology direct  \textbf{5}, ~1--9 (2010)

\bibitem{esmeral2022low}
Esmeral, L.C.M., Uhl, A.: Low-effort re-identification techniques based on medical imagery threaten patient privacy. In: Annual Conference on Medical Image Understanding and Analysis. pp. 719--733. Springer (2022)

\bibitem{gdpr}
{European Union}: Complete guide to gdpr compliance (2023), \url{https://gdpr.eu/}, accessed: 08 Aug 2023

\bibitem{ganz2021}
Ganz, J., Kirsch, T., Hoffmann, L., Bertram, C.A., Hoffmann, C., Maier, A., Breininger, K., Bl{\"u}mcke, I., Jabari, S., Aubreville, M.: Automatic and explainable grading of meningiomas from histopathology images. In: MICCAI Workshop on Computational Pathology. pp. 69--80. PMLR (2021)

\bibitem{han2017breast}
Han, Z., Wei, B., Zheng, Y., Yin, Y., Li, K., Li, S.: Breast cancer multi-classification from histopathological images with structured deep learning model. Scientific reports  \textbf{7}(1), ~4172 (2017)

\bibitem{hong2021predicting}
Hong, R., Liu, W., DeLair, D., Razavian, N., Feny{\"o}, D.: Predicting endometrial cancer subtypes and molecular features from histopathology images using multi-resolution deep learning models. Cell Reports Medicine  \textbf{2}(9),  100400 (2021)

\bibitem{Howard2021}
Howard, F.M., Dolezal, J., Kochanny, S., Schulte, J., Chen, H., Heij, L., Huo, D., Nanda, R., Olopade, O.I., Kather, J.N., et~al.: The impact of site-specific digital histology signatures on deep learning model accuracy and bias. Nature communications  \textbf{12}(1), ~4423 (2021)

\bibitem{chest1}
Ishigami, R., Zin, T.T., Shinkawa, N., Nishii, R.: Human identification using x-ray image matching. In: Proceedings of the International Multi Conference of Engineers and Computer Scientists. vol.~1 (2017)

\bibitem{janowczyk2019histoqc}
Janowczyk, A., Zuo, R., Gilmore, H., Feldman, M., Madabhushi, A.: Histoqc: an open-source quality control tool for digital pathology slides. JCO clinical cancer informatics  \textbf{3}, ~1--7 (2019)

\bibitem{hand1}
Kabbara, Y., Shahin, A., Nait-Ali, A., Khalil, M.: An automatic algorithm for human identification using hand x-ray images. In: 2013 2nd International Conference on Advances in Biomedical Engineering. pp. 167--170. IEEE (2013)

\bibitem{keller2023tissue}
Keller, P., Dawood, M., Minhas, F.u.A.: Do tissue source sites leave identifiable signatures in whole slide images beyond staining? In: International Workshop on Trustworthy Machine Learning for Healthcare. pp. 1--10. Springer (2023)

\bibitem{van2003tumor}
van Kempen, L.C., Ruiter, D.J., van Muijen, G.N., Coussens, L.M.: The tumor microenvironment: a critical determinant of neoplastic evolution. European journal of cell biology  \textbf{82}(11),  539--548 (2003)

\bibitem{brain1}
Kumar, K., Desrosiers, C., Siddiqi, K., Colliot, O., Toews, M.: Fiberprint: A subject fingerprint based on sparse code pooling for white matter fiber analysis. NeuroImage  \textbf{158},  242--259 (2017)

\bibitem{laconi2007evolving}
Laconi, E.: The evolving concept of tumor microenvironments. Bioessays  \textbf{29}(8),  738--744 (2007)

\bibitem{TCGA-BRCA}
Lingle, W., Erickson, B.J., Zuley, M.L., Jarosz, R., Bonaccio, E., Filippini, J., Net, J.M., Levi, L., Morris, E.A., Figler, G.G., Elnajjar, P., Kirk, S., Lee, Y., Giger, M., Gruszauskas, N.: The cancer genome atlas breast invasive carcinoma collection (tcga-brca) (version 3) [dataset] (2016)

\bibitem{litjens20181399}
Litjens, G., Bandi, P., Ehteshami~Bejnordi, B., Geessink, O., Balkenhol, M., Bult, P., Halilovic, A., Hermsen, M., van~de Loo, R., Vogels, R., et~al.: 1399 h\&e-stained sentinel lymph node sections of breast cancer patients: the camelyon dataset. GigaScience  \textbf{7}(6),  giy065 (2018)

\bibitem{iris2}
Liu, N., Zhang, M., Li, H., Sun, Z., Tan, T.: Deepiris: Learning pairwise filter bank for heterogeneous iris verification. Pattern Recognition Letters  \textbf{82},  154--161 (2016)

\bibitem{lu2021cup}
Lu, M.Y., Chen, T.Y., Williamson, D.F., Zhao, M., Shady, M., Lipkova, J., Mahmood, F.: Ai-based pathology predicts origins for cancers of unknown primary. Nature  \textbf{594}(7861),  106--110 (2021)

\bibitem{lu2021clam}
Lu, M.Y., Williamson, D.F., Chen, T.Y., Chen, R.J., Barbieri, M., Mahmood, F.: Data-efficient and weakly supervised computational pathology on whole-slide images. Nature Biomedical Engineering  \textbf{5}(6),  555--570 (2021)

\bibitem{macenko2009method}
Macenko, M., Niethammer, M., Marron, J.S., Borland, D., Woosley, J.T., Guan, X., Schmitt, C., Thomas, N.E.: A method for normalizing histology slides for quantitative analysis. In: 2009 IEEE international symposium on biomedical imaging: from nano to macro. pp. 1107--1110. IEEE (2009)

\bibitem{palm2}
Matkowski, W.M., Chai, T., Kong, A.W.K.: Palmprint recognition in uncontrolled and uncooperative environment. IEEE Transactions on Information Forensics and Security  \textbf{15},  1601--1615 (2019)

\bibitem{moore2015identification}
Moore, S.M., Maffitt, D.R., Smith, K.E., Kirby, J.S., Clark, K.W., Freymann, J.B., Vendt, B.A., Tarbox, L.R., Prior, F.W.: De-identification of medical images with retention of scientific research value. Radiographics  \textbf{35}(3),  727--735 (2015)

\bibitem{LUAD}
{National Cancer Institute Clinical Proteomic Tumor Analysis Consortium (CPTAC)}: The clinical proteomic tumor analysis consortium lung adenocarcinoma collection (cptac-luad) (version 12)[dataset] (2018)

\bibitem{LSCC}
{National Cancer Institute Clinical Proteomic Tumor Analysis Consortium (CPTAC)}: The clinical proteomic tumor analysis consortium lung squamous cell carcinoma collection (cptac-lscc) (version 14) [dataset] (2018)

\bibitem{NIR2018167}
Nir, G., Hor, S., Karimi, D., Fazli, L., Skinnider, B.F., Tavassoli, P., Turbin, D., Villamil, C.F., Wang, G., Wilson, R.S., Iczkowski, K.A., Lucia, M.S., Black, P.C., Abolmaesumi, P., Goldenberg, S.L., Salcudean, S.E.: Automatic grading of prostate cancer in digitized histopathology images: Learning from multiple experts. Medical Image Analysis  \textbf{50},  167--180 (2018)

\bibitem{teeth1}
Nomir, O., Abdel-Mottaleb, M.: Human identification from dental x-ray images based on the shape and appearance of the teeth. IEEE transactions on information forensics and security  \textbf{2}(2),  188--197 (2007)

\bibitem{otalora2022stainlib}
Ot{\'a}lora, S., Marini, N., Podareanu, D., Hekster, R., Tellez, D., van~der Laak, J., M{\"u}ller, H., Atzori, M.: Stainlib: a python library for augmentation and normalization of histopathology h\&e images. bioRxiv pp. 2022--05 (2022)

\bibitem{otsu1979threshold}
Otsu, N.: A threshold selection method from gray-level histograms. IEEE transactions on systems, man, and cybernetics  \textbf{9}(1),  62--66 (1979)

\bibitem{packhauser2022deep}
Packh{\"a}user, K., G{\"u}ndel, S., M{\"u}nster, N., Syben, C., Christlein, V., Maier, A.: Deep learning-based patient re-identification is able to exploit the biometric nature of medical chest x-ray data. Scientific Reports  \textbf{12}(1),  14851 (2022)

\bibitem{face1}
Parkhi, O., Vedaldi, A., Zisserman, A.: Deep face recognition. In: BMVC 2015 - Proceedings of the British Machine Vision Conference. pp. 1--12. British Machine Vision Association (2015)

\bibitem{knee2}
Shamir, L.: Mri-based knee image for personal identification. International Journal of Biometrics  \textbf{5}(2),  113--125 (2013)

\bibitem{knee1}
Shamir, L., Ling, S., Rahimi, S., Ferrucci, L., Goldberg, I.G.: Biometric identification using knee x-rays. International journal of biometrics  \textbf{1}(3),  365--370 (2009)

\bibitem{sweeney2002k}
Sweeney, L.: k-anonymity: A model for protecting privacy. International journal of uncertainty, fuzziness and knowledge-based systems  \textbf{10}(05),  557--570 (2002)

\bibitem{veta2019predicting}
Veta, M., Heng, Y.J., Stathonikos, N., Bejnordi, B.E., Beca, F., Wollmann, T., Rohr, K., Shah, M.A., Wang, D., Rousson, M., et~al.: Predicting breast tumor proliferation from whole-slide images: the tupac16 challenge. Medical image analysis  \textbf{54},  111--121 (2019)

\bibitem{willemink2020preparing}
Willemink, M.J., Koszek, W.A., Hardell, C., Wu, J., Fleischmann, D., Harvey, H., Folio, L.R., Summers, R.M., Rubin, D.L., Lungren, M.P.: Preparing medical imaging data for machine learning. Radiology  \textbf{295}(1),  4--15 (2020)

\bibitem{Wilm2022}
Wilm, F., Fragoso, M., Marzahl, C., Qiu, J., Puget, C., Diehl, L., Bertram, C.A., Klopfleisch, R., Maier, A., Breininger, K., et~al.: Pan-tumor canine cutaneous cancer histology (catch) dataset. Scientific Data  \textbf{9}(1), ~588 (2022)

\bibitem{finger1}
Zeng, F., Hu, S., Xiao, K.: Research on partial fingerprint recognition algorithm based on deep learning. Neural Computing and Applications  \textbf{31}(9),  4789--4798 (2019)

\bibitem{iris1}
Zhao, T., Liu, Y., Huo, G., Zhu, X.: A deep learning iris recognition method based on capsule network architecture. IEEE Access  \textbf{7},  49691--49701 (2019)

\bibitem{palm1}
Zhong, D., Du, X., Zhong, K.: Decade progress of palmprint recognition: A brief survey. Neurocomputing  \textbf{328},  16--28 (2019), chinese Conference on Computer Vision 2017

\bibitem{teeth2}
Zhou, J., Abdel-Mottaleb, M.: A content-based system for human identification based on bitewing dental x-ray images. Pattern Recognition  \textbf{38}(11),  2132--2142 (2005)

\end{thebibliography}
\end{document}

\verso{Jonathan Ganz \textit{et~al.}}
\appendix
\setcounter{table}{0}
\section{Investigation on the optimal magnification level for sampling}
\label{appx:levels}
\begin{table}[ht]
        \caption{Results of the preliminary investigation of the optimal magnification level for patch sampling. Given are the results of the tenfold Monte Carlo cross-validation using the \ac{MEN} dataset and the patch-based model. In each experiment, patches with a width and height of 512 pixels were used. The spatial resolution is given in microns per pixel (mpp).}
        \resizebox{\linewidth}{!}{
        \centering
        \begin{tabular}{lcccccccc}
        \hline
        Spatial Resolution & recall@1 & recall@5 & precision & $F_1$ core\\
        \hline
        $0.22$ mpp & $54.79\% \pm 5.10\%$ & $73.08\% \pm 4.23\%$ & $52.14\% \pm 5.92\%$ & $51.24\% \pm 5.47\%$ \\
        $0.44$ mpp & $56.43\% \pm 5.23\%$ & $75.37\% \pm 4.22\%$ & $54.65\% \pm 6.82\%$ & $53.05\% \pm 6.08\%$ \\
        $0.88$ mpp & $65.25\% \pm 6.49\%$ & $80.59\% \pm 4.02\%$ & $63.79\% \pm 6.30\%$ & $62.31\% \pm 6.45\%$ \\
        $1.76$ mpp & $64.95\% \pm 5.45\%$ & $80.63\% \pm 3.90\%$ & $63.54\% \pm 6.40\%$ & $62.03\% \pm 5.91\%$ \\
        $3.52$ mpp & $63.75\% \pm 6.07\%$ & $78.60\% \pm 4.22\%$ & $61.90\% \pm 6.24\%$ & $60.52\% \pm 6.16\%$ \\
        $7.04$ mpp & $59.16\% \pm 3.78\%$ & $73.29\% \pm 3.40\%$ & $57.09\% \pm 3.95\%$ & $55.89\% \pm 3.81\%$ \\
        \hline
        \end{tabular}}
    \label{tab:levels}
\end{table}
 In a preliminary experiment, we investigated the most effective magnification level for training the re-identification models. We achieved this by conducting a tenfold Monte Carlo cross-validation on six varying magnification levels utilizing our patch-based classification model on the MEN dataset. At the highest magnification level, our spatial resolution was 0.22 \ac{mpp}, which halved on each subsequent level. See Table \ref{tab:levels} for the comprehensive experiment results. We discovered that performance continued to improve until a spatial resolution of 0.88 \ac{mpp}. At lower magnifications, however, performance degraded. We consider a resolution of 0.88 \ac{mpp} to be an adequate compromise between microscopic features such as the form of various nuclei and macroscopic features like cell formation. We therefore used a spatial resolution of 0.88 \ac{mpp} in all experiments with all datasets in this study.